\documentclass[conference]{IEEEtran}
\IEEEoverridecommandlockouts
\usepackage{cite}
\usepackage{amsmath,amssymb,amsfonts}
\usepackage{algorithmic,algorithm}
\usepackage{graphicx}
\usepackage{textcomp}
\usepackage{xcolor}

\usepackage{hyperref}
\usepackage{amsmath}
\usepackage{comment}

\DeclareMathOperator*{\argmin}{arg\,min}

\def\BibTeX{{\rm B\kern-.05em{\sc i\kern-.025em b}\kern-.08em
    T\kern-.1667em\lower.7ex\hbox{E}\kern-.125emX}}
\begin{document}

\title{Towards adversarial robustness with 01 loss neural networks\\
}

\author{\IEEEauthorblockN{1\textsuperscript{st} Yunzhe Xue}
\IEEEauthorblockA{\textit{Department of Computer Science} \\
\textit{New Jersey Institute of Technology}\\
Newark, USA \\
yx277@njit.edu}
\and
\IEEEauthorblockN{2\textsuperscript{nd} Meiyan Xie}
\IEEEauthorblockA{\textit{Department of Computer Science} \\
\textit{New Jersey Institute of Technology}\\
Newark, USA\\
mx42@njit.edu}
\and
\IEEEauthorblockN{3\textsuperscript{rd} Usman Roshan}
\IEEEauthorblockA{\textit{Department of Computer Science} \\
\textit{New Jersey Institute of Technology}\\
Newark, USA\\
usman@njit.edu}
}

\maketitle

\begin{abstract}
Motivated by the general robustness properties of the 01 loss we propose a single hidden layer 01 loss neural network trained with stochastic coordinate descent as a defense against adversarial attacks in machine learning. One measure of a model's robustness is the minimum distortion required to make the input adversarial. This can be approximated with the Boundary Attack (Brendel \emph{et. al.} 2018) and HopSkipJump (Chen \emph{et. al.} 2019) methods. We compare the minimum distortion of the 01 loss network to the binarized neural network and the standard sigmoid activation network with cross-entropy loss all trained with and without Gaussian noise on the CIFAR10 benchmark binary classification between classes 0 and 1. Both with and without noise training we find our 01 loss network to have the largest adversarial distortion of the three models by non-trivial margins. To further validate these results we subject all models to substitute model black box attacks under different distortion thresholds and find that the 01 loss network is the hardest to attack across all distortions. At a distortion of 0.125 both sigmoid activated cross-entropy loss and binarized networks have almost 0\% accuracy on adversarial examples whereas the 01 loss network is at 40\%. Even though both 01 loss and the binarized network use sign activations their training algorithms are different which in turn give different solutions for robustness. Finally we compare our network to simple convolutional models under substitute model black box attacks and find their accuracies to be comparable. Our work shows that the 01 loss network has the potential to defend against black box adversarial attacks better than convex loss and binarized networks. 
\end{abstract}


\section{Introduction}
State of the art machine learning algorithms can achieve high accuracies in classification tasks but misclassify minor perturbations in the data known as as adversarial attacks 
\cite{goodfellow2014explaining,papernot2016limitations,kurakin2016adversarial,carlini2017towards,brendel2017decision}. Adversarial examples have been shown to transfer across models which makes it possible to perform transfer-based (substitute model) black box attacks \cite{papernot2016transferability}. To counter adversarial attacks many defense methods been proposed with adversarial training being the most popular \cite{szegedy2013intriguing}. This is known to improve robustness to adversarial examples but also tends to lower accuracy on clean test data that has no perturbations \cite{raghunathan2019adversarial,zhang2019theoretically}. Many previously proposed defenses have also shown to be vulnerable \cite{carlini2017towards,athalye2018obfuscated,ghiasi2020breaking} thus leaving adversarial robustness an open problem in machine learning.

The 01 loss is known to be more robust to outliers than convex loss models \cite{xie2019,icml13optimize,bartlett04}. In addition to being robust to outliers the 01 loss is also robust to noise in the training data \cite{manwani2013noise,ghosh2015making}. Under this loss minimizing the empirical risk amounts to minimizing the empirical adversarial risk \cite{lyu2019curriculum,hu2016does} with certain assumptions of noise. Convex losses also fail to minimize the adversarial 01 loss on linear models \cite{bao2020calibrated}. 

Motivated by the above robustness properties of 01 loss we propose a 01 loss dual layer neural network as a defense against adversarial attacks. Computationally 01 loss presents a considerable challenge because it is NP-hard to solve \cite{ben03}. Previous attempts \cite{nips01optimize,mixedint01,approx01,ijcnn01optimize,icml13optimize} lack on-par test accuracy with convex solvers and are slow and impractical for large image benchmarks. However, a recent stochastic coordinate descent method for linear 01 loss models \cite{xie2019} has shown to attain comparable accuracies to state of the art linear solvers like the support vector machine. Thus we extend the coordinate descent as an optimizer to train our network. 

We compare the adversarial robustness of our model to an equivalent one that uses sigmoid activation and cross-entropy loss. This is the standard activation and loss that are widely used in neural networks today. We also compare our model to the binarized neural network  \cite{galloway2017attacking,courbariaux2016binarized,rastegari2016xnor}. that also uses sign activations like our model but it has two differences. First its weights are also constrained to be binary +1 and -1, or 1 and 0 \cite{galloway2017attacking,courbariaux2016binarized,rastegari2016xnor}. Second, it is trained with gradient descent by approximating the sign activation whereas we take a direct coordinate descent approach.

Measuring adversarial robustness is not trivial and several best practices have been recommended \cite{carlini2019evaluating}. We incorporate several of them in our study. In particular we study (1) the robustness of models to random Gaussian noise, (2) the minimum distortion required to make a datapoint adversarial, and (3) the accuracy of substitute model black box attacks. We focus mainly on binary classification between classes 0 and 1 on the CIFAR10 image benchmark \cite{krizhevsky2009learning} where we make the following findings. 

\begin{itemize}
\item All models are more robust to Gaussian noise than adversarial attacks, but our 01 loss network augmented with Gaussian noise during training has a higher accuracy on large distortions
\item The minimum distortion to make an image adversarial is higher for our 01 loss network compared to the standard sigmoid activated cross-entropy loss network and binarized networks
\item Substitute model black box attacks are far less effective on our 01 loss network compared to the standard sigmoid activated cross-entropy loss network and binarized networks
\item Compared to simple convolutional neural networks like LeNet \cite{lecun1998gradient} our model (without convolutions) has higher accuracies on adversarial examples from substitute model black box attacks when the distortions are high. 
\end{itemize}

\section{Methods}

\subsection{Background}
The problem of determining the hyperplane with minimum number of misclassifications
in a binary classification problem is known to be NP-hard \cite{ben03}.
In mainstream machine learning literature this is called minimizing the 01 loss
\cite{kernel01} given in Objective~\ref{obj1},

\begin{equation}
\frac{1}{2n}\argmin_{w,w_0} \sum_i (1-sign(y_i(w^Tx_i+w_0)))
\label{obj1}
\end{equation}

where $w \in R^d$, $w_0 \in R$ is our hyperplane, and $x_i \in R^d, y_i\in \{+1,-1\}.\forall i=0...n-1$ are our training data. Popular linear classifiers such as the linear support 
vector machine, perceptron, and logistic regression \cite{alpaydin} can be considered 
as convex approximations to this problem that yield fast gradient descent solutions \cite{bartlett04}. 
However, they are also more sensitive to outliers than the 01 loss \cite{bartlett04,icml13optimize,xie2019}
and more prone to mislabeled data than 01 loss \cite{manwani2013noise,ghosh2015making,lyu2019curriculum}. 

\subsection{A dual layer 01 loss neural network}
We extend the 01 loss to a simple two layer neural network with $k$ hidden nodes and sign activation that we call the MLP01 loss. This objective for binary classification can be given as

\begin{equation}
\small
\frac{1}{2n}\argmin_{W, W_0, w,w_0} \sum_i (1-sign(y_i(w^T(sign(W^Tx_i+W_0))+w_0)))
\label{obj2}
\end{equation}

where $W \in R^{d\times k}$, $W_0 \in R^k$ are the hidden layer parameters, $w\in R^k, w_0\in R$ are the final layer node parameters, $x_i \in R^d, y_i\in \{+1,-1\}.\forall i=0...n-1$ are our training data, and $sign(v\in R^k)=(sign(v_0), sign(v_1),...,sign(v_{k-1}))$. While this is a straightforward model to define optimizing it is a different story altogether. Optimizing even a single node is NP-hard which makes optimizing this network much harder. 



\subsection{Stochastic coordinate descent for 01 loss}

In Algorithm~\ref{mlp01} we sketch our coordinate descent for our 01 loss network that is based upon earlier work \cite{xie2019}. We initialize all parameters to random values from the Normal distribution with mean 0 and variance 1. We then randomly select a subset of the training data (known as a batch) and perform the coordinate descent analog of a single step gradient update in stochastic gradient descent \cite{bottou2010large}.

\begin{algorithm}[!h]
\caption{Stochastic coordinate descent for two layer 01 loss network} 
\label{mlp01}
\textbf{Procedure: }
\begin{algorithmic}
\STATE 1. Initialize all network weights $W,w$ to random values from the Normal distribution $N(0,1)$.
\STATE 2. Set network thresholds $W_0$ to the median projection value on their corresponding weight vectors and $w_0$ to the projection value that minimizes our network objective.
\WHILE {$i < epochs$} 
	\STATE 1. Randomly sample a batch of data equally from each class
	\STATE 2. Perform coordinate descent separately first on the final node $w$ and then a randomly selected hidden node $u$ (a random column from the hidden layer weight matrix $W$)
	\STATE 3. In the coordinate descent we randomly pick a set of features and perform a single update to each one. For each update we determine the threshold that optimizes the 01 loss on the sampled data: we sort the projections $w^Tx_i$ and pick the optimal middle value between each consecutive pair in the projected values. 
	\STATE 4. After making the best update we evaluate the 01 loss on the full dataset and accept the change if it improves the loss. 
\ENDWHILE
\end{algorithmic}
\end{algorithm}

When the gradient is known we step in its negative direction by a factor of the learning rate: $w=w-\eta\nabla(f)$ where $f$ is the objective. In our case since the gradient does not exist we randomly select $k$ features (set to 128 in our experiments), modify the corresponding entries in $w$ by the learning rate (set to 0.17) one at a time, and accept the modification that gives the largest decrease in the objective. Key to our search is a heuristic to determine the optimal threshold each time we modify an entry of $w$. In this heuristic we perform a linear search on a subset of the projection $w^Tx_i$ and select $w_0$ that minimizes the objective.

\begin{figure}[h]
  \centering
  \includegraphics[trim=80 50 0 70, clip, scale=.27]{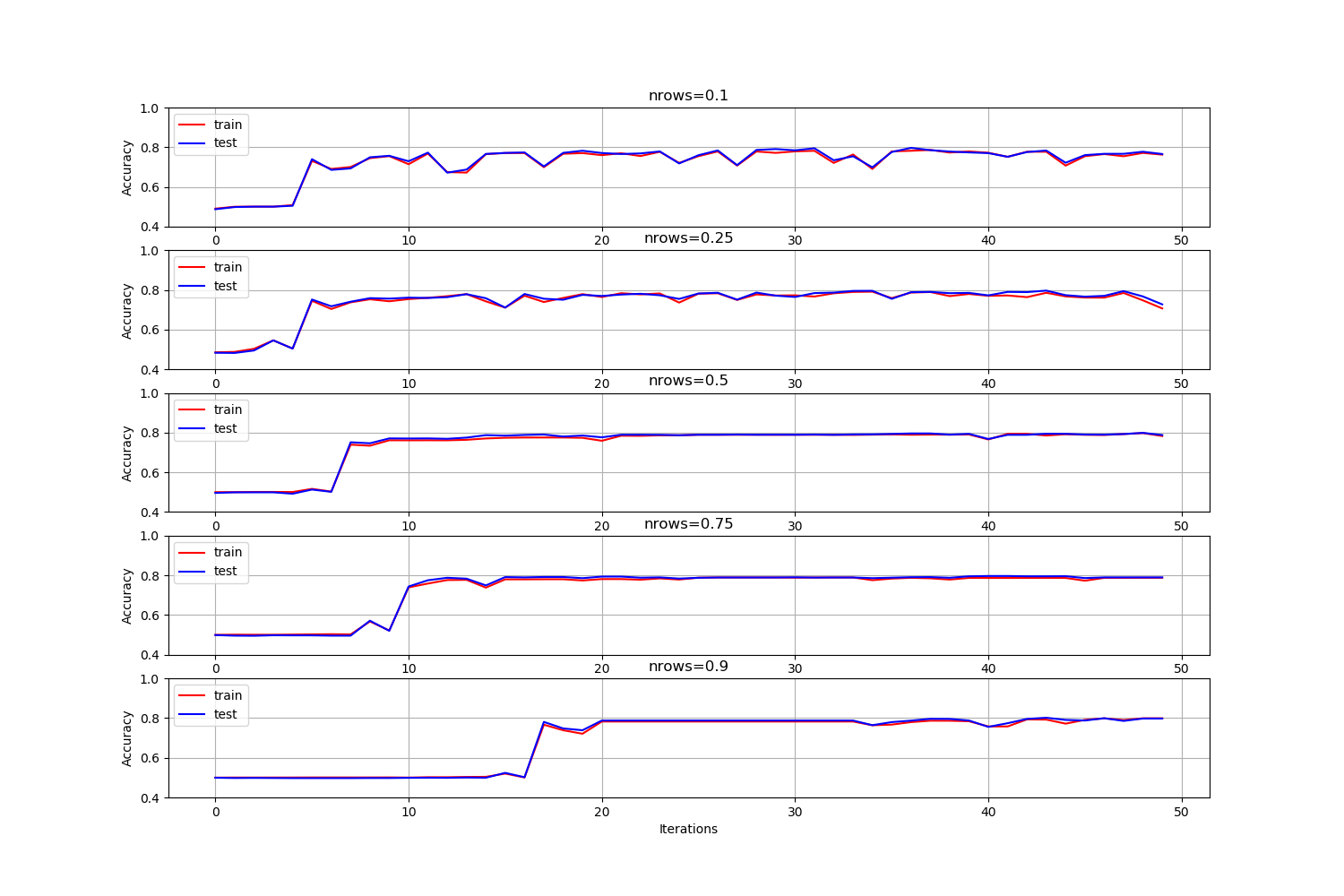} 
  \caption{Train and test accuracy of our stochastic coordinate descent on CIFAR10 class 0 vs 1 with different batch sizes (denoted as nrows). \label{nrows}}
\end{figure}

In Figure~\ref{nrows} we show the effect of the batch size (as a percentage of each class to ensure fair sampling) on a linear 01 loss search on CIFAR10 between classes 0 and 1. We see that a batch size of 75\% reaches a train accuracy of 80\% faster than the other batch sizes. Thus we use this batch size in all our experiments going forward.

We also see that for this batch size the search flattens after 15 iterations (or epochs as given in the figure). We run 1000 iterations to ensure a deep search with an intent to maximize test accuracy. The problem with our search described above is that it will return different solutions depending upon the initial starting point. To make it more stable we run it a 100 times from different random seeds and use the majority vote for prediction. Full details of our training algorithms are in the Supplementary Material.

\subsection{Implementation, experimental platform, and image data}
\subsubsection{Implementation}
We implement our 01 loss network (MLP01) in Python and Pytorch \cite{pytorch}, the sigmoid activated cross-entropy loss network (MLP) in scikit-learn \cite{scikit}, and binarized neural network (BNN) with the Larq software suite \url{https://github.com/larq/larq}. We train MLP with stochastic gradient descent that has a batch size of 200, momentum of 0.9, and learning rate of 0.01. For BNN we use the approximate sign activation \cite{courbariaux2016binarized} that has been shown to give higher test accuracies than other variants and the original straight through estimator \cite{liu2018bi}.

\subsubsection{Computational platform}
We ran all experiments on Intel Xeon 6142 2.6GHz CPUs and NVIDIA Titan RTX GPU machines (for parallelizing multiple votes). Our MLP01 source code, supplementary programs, and data are available from \url{https://github.com/zero-one-loss/mlp01}. 

\subsubsection{Data}
We experiment on the popular image benchmark CIFAR10 \cite{krizhevsky2009learning}
that has $32\times32$ color images with 50000 training and 10000 test. We extract data from classes 0 and 1 and experiment on binary classification between them. This gives us a total of 10000 training and 1000 test examples. We normalize each image by dividing each pixel value by 255.

\section{Results}
We refer to our 01 loss neural network as MLP01, the sigmoid activated cross-entropy loss network as MLP, and the binarized network as BNN. We use one hidden layer of 20 nodes in all networks. For each model we run it a 100 times with different random number generator seeds and return the majority vote as the prediction. 

In addition to training each model on the training data, we also study three versions trained with augment Gaussian noise. In the augmentation we take each datapoint $x$ from the training set and add Gaussian noise to it: $x'=x+N(0,\sigma)$ where $N(0,\sigma)$ is a vector of the same dimension as $x$ and each entry is selected from the Normal distribution with mean 0 and standard deviation $\sigma$. 

We refer to the accuracy on the test data as clean data test accuracy. An incorrectly classified adversarial example is considered a successful attack whereas a correctly classified adversarial is a failed one. Thus when we refer to accuracy of adversarial examples it is the same as $100-attack success rate$. The lower the accuracy the more effective the attack.

\subsection{Sensitivity to Gaussian noise}
We start with accuracy of models trained without and with noise on CIFAR10 class 0 vs. 1. In addition to evaluating the accuracy of each model on clean test data, we add noise to each test datapoint as $x'=x+N(0,\sigma)$ where $N(0,\sigma)$ is a vector of the same dimension as $x$ and each entry is selected from the Normal distribution with mean 0 and standard deviation $\sigma$. We consider $\sigma$ ranging from .004 to 1. The lower bound is the minimum distance of $\frac{1}{255}$ between two pixels and $1$ is the maximum distortion. If a pixel is negative or above 1 after adding noise we clip it to 0 and 1 respectively.

\begin{figure}[!h]
  \centering
  \includegraphics[scale=.425]{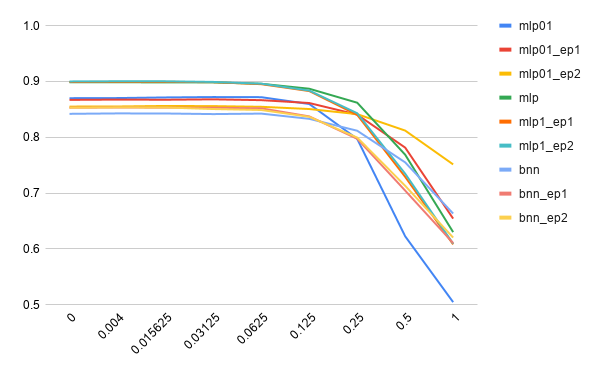} 
  \caption{Accuracy of test data without and with random noise of different random Gaussian distortions. In addition to models trained on clean training data we consider two versions trained with augmented Gaussian noise with distortion thresholds of 0.1 and 0.2 (denoted with ep1 and ep2 respectivelt). \label{noise}}
\end{figure}

In Figure~\ref{noise} we see that noise does not affect the accuracy of models trained without and with noise upto distortion threshold of 0.125. After that all models begin to dip in accuracy with MLP01 model trained without noise showing the steepest descent. At the same time MLP01 trained with noisy augmentation of $\sigma=.2$ (denoted as mlp01\_ep2 in Figure~\ref{noise}) is also most robust to high levels of noise.

\subsection{Minimum adversarial distortion}
Determining the minimum distortion to an image such that it will fool a classifier is itself an NP-hard problem for ReLu activated neural networks \cite{katz2017reluplex,sinha2017certifiable} and tree ensemble classifiers \cite{kantchelian2016evasion}. Even approximating the minimum distortion for ReLu activated neural networks is NP-hard \cite{weng2018towards}. Recent heuristics such as Boundary attack \cite{brendel2017decision} and HopSkipJump attack \cite{chen2019hopskipjumpattack} attempt to find an adversarial example with the minimum distance to the clean correctly classified version. 

We use both methods to evaluate the minimum distortion of all three models including their noise trained versions. Both methods can take long to finish with default parameters even for a single example. For example for a single image BNN takes 3 days on an exclusive CPU core. Thus we use all default parameters except for maxiter that we set to 100 so that the program finishes within our computing limitations. To confirm that this does not severely affect the relative distortions we ran both Boundary and HopSkipJump attacks with maxiter set to 10, 100, and 500 (which is the default) on a single image. We found the relative distortions between MLP and MLP01 to be the same across the three values.

Since both attack methods start with a random initialization we run each of them 10 times for a single example on each model and report the minimum value found. In Table~\ref{distortion} we report these values for a single random test datapoint from CIFAR10 classes 0 and 1 that is correctly classified by all models. We see that the minimum distortion of MLP01 is much higher than both MLP and BNN by both attack methods and under both $L_2$ and $L_\infty$ norms. We also see that HopSkipJump attack is more effective than Boundary attack and finds a smaller distortion. 

\begin{table}[!h]
  \caption{Minimum adversarial distortion of a single random test image \label{distortion}}
  \centering
  \begin{tabular}{lllllll} \hline
   & \multicolumn{3}{c}{$L_2$ distance} &  \multicolumn{3}{c}{$L_{\infty}$ distance} \\
                    & BNN  & MLP & MLP01 & BNN  & MLP & MLP01  \\
    Boundary &  2.72  & 1.33 & 10.58  & 0.17 & 0.08 & 0.58 \\
    HopSkipJump & 0.82 & 0.44 & 2.21 & 0.04 & 0.03 & 0.16 \\
  \end{tabular}
\end{table}

In Table~\ref{distortion2} we report the HopSkipJump distortions for four more randomly selected images from CIFAR10 classes 0 and 1 that are correctly classified. For the first image both have comparable distortion but for the other three MLP01 is higher. 

\begin{table}[!h]
  \caption{Minimum adversarial distortion given by HopSkipJump of four random correctly classified test images \label{distortion2}}
  \centering
  \begin{tabular}{llllll} \hline
   & \multicolumn{2}{c}{$L_2$ distance} &  \multicolumn{2}{c}{$L_{\infty}$ distance} \\
                  & MLP & MLP01 & MLP & MLP01 \\
    Image1 &   .64 & .52 & .043 & .041 \\
    Image2 &   .75 & 2.42 & .06 & .15  \\
    Image3 &   .88 & 1.12 & .06 & .09 \\
    Image4 &   1.15 & 4.86 & .09 & .28 \\ \hline
    Average & .86 & 2.23 & .063 &  0.14 \\
  \end{tabular}
\end{table}

Training our models with noise has an interesting effect on the distortions. In Table~\ref{distortion3} we see the minimum distortions of models trained with Gaussian noise with distortions of 0.1 and 0.2 (as described earlier). We report the distortions for the same image as in Table~\ref{distortion}. As we increase the noise threshold the MLP01 model's minimum distortion also rises whereas the other two models are stable or fluctuate. 

\begin{table}[!h]
  \caption{Minimum adversarial distortion of the single test image from Table~\ref{distortion} as given by HopSkipJump. Models are trained with Gaussian noise of increasing distortion shown by $\epsilon$. \label{distortion3}}
  \centering
  \begin{tabular}{lllllll} \hline
   & \multicolumn{3}{c}{$L_2$ distance} &  \multicolumn{3}{c}{$L_{\infty}$ distance} \\
                    & $\epsilon=.004$  &$\epsilon=.1$ & $\epsilon=.2$ & $\epsilon=.004$  &$\epsilon=.1$ & $\epsilon=.2$  \\
    BNN &  .51 & .47 & .53 & .025 & .021 & .025 \\
    MLP &   .39 & .35 & .36 & .023 & .021 & .022 \\
    MLP01    & 2.52 & 2.61 & 3.32 & .196 & .215 & .22 \\
  \end{tabular}
\end{table}

We make similar observations between MLP and MLP01 on four random examples as shown in Table~\ref{distortion4}. This suggests that perhaps training MLP01 with augmented noise examples increases their robustness.

\begin{table}[!h]
  \caption{Minimum adversarial distortion of four random correctly classified test images from Table~\ref{distortion2} as given by HopSkipJump. Models are trained with Gaussian noise of increasing distortion shown by $\epsilon$. \label{distortion4}}
  \centering
  \begin{tabular}{lllllll} \hline
     & \multicolumn{6}{c}{$L_2$ distance}  \\

   & \multicolumn{3}{c}{MLP} &  \multicolumn{3}{c}{MLP01} \\
                    & $\epsilon=.004$  &$\epsilon=.1$ & $\epsilon=.2$ & $\epsilon=.004$  &$\epsilon=.1$ & $\epsilon=.2$  \\
    Image1 & .58 & .47 & .47 & .69 & .61 & .47  \\
    Image2  & .6 & .52 & .54 &   2.65 & 2.89 & 2.51\\
    Image3   & .72 & .67 & .66  & 1.92 & 2.72 & 4.03\\
    Image4  & .97 & .85 & .84   & 4.73 & 4.08 & 5.17 \\ \hline
    Average & .72 & 63 & .63  & 2.5  & 2.58 & 3.05  \\ \hline \hline
     & \multicolumn{6}{c}{$L_\infty$ distance}  \\

   & \multicolumn{3}{c}{MLP} &  \multicolumn{3}{c}{MLP01} \\
                    & $\epsilon=.004$  &$\epsilon=.1$ & $\epsilon=.2$ & $\epsilon=.004$  &$\epsilon=.1$ & $\epsilon=.2$  \\
    Image1 & .04 & .03 & .03  &  .05 & .05 & .03 \\
    Image2 & .04 & .03 & .03 &   .18 & .2 & .18 \\
    Image3   & .05 & .04 & .04 & .14 & .21 & .35 \\
    Image4   & .07 & .05 & .05  & .28 & .29 & .35 \\ \hline
    Average &  .05 & .04 & .04 & .16 & .19 & .23  \\ \hline 

  \end{tabular}
\end{table}


\subsection{Substitute model black box attacks}
As further verification of the above distortions we perform substitute model black box attacks on all three models. In this method we try to approximate the target model with a substitute and then generate white box adversaries from the substitute to attack the target model. The success of this method relies upon transferability of adversarial examples between models. We use the standard adversarially augmented training algorithm of Papernot et. al. \cite{papernot2017practical} to train the substitute. In the Supplementary Material we provide full details of the algorithm.

This in fact is a powerful attack method that needs only predicted labels from the target (like Boundary and HopSkipJump) but requires much fewer queries. Once the substitute is trained it can produce adversaries for any input. Recent advances in transferability have made this method more effective and broken defenses based on adversarial training \cite{tramer2017ensemble,wu2020skip}. For the substitute model we use a three layer sigmoid activated cross-entropy loss network with 200 nodes in each hidden layer. We start with 200 random test data points from which we iteratively train the substitute model with augmented adversaries. 

\begin{figure}[!h]
  \centering
  \includegraphics[scale=.425]{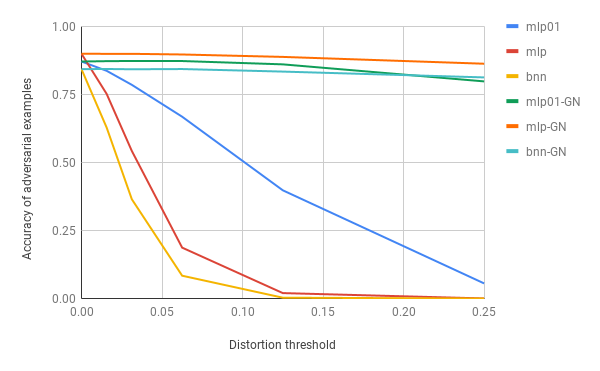} 
  \caption{Accuracy of adversarial examples and test data with Gaussian noise (denoted as -GN for each model) for various distortion thresholds. The adversarial examples are far more effective than random noise. At distortion 0.125 both MLP and BNN have near 0\% accuracy whereas MLP01 has 40\%. \label{blackbox}}
\end{figure}

In Figure~\ref{blackbox} we see the accuracy of adversarial examples at the end of the $20^{th}$ epoch. We also show the accuracy of the three models on random Gaussian noise of the same distortions (from our earlier subsection above). Clearly the black box adversaries are far more effective than random noise indicating that the substitute model training was successful. In agreement with our distortions from Boundary and HopSkipJump above we see that MLP01 can correctly classify images of much higher distortion than BNN and MLP. 

\subsection{Comparison to convolutional neural networks}
As a test against state of the art classification methods we compare our 01 loss network with 500 hidden nodes to two convolutional neural networks. First is LeNet \cite{lecun1998gradient} which is among the first convolutional networks to be proposed and second is SimpleNet500. In this model we use the same convolutional layers as LeNet followed by one layer of 500 nodes and then the final output node. 

We employ the same substitute model training algorithm as in the above subsection. However instead of a dual hidden layer model we use a convolutional network as the substitute. In each convolutional block we have a $3\times3$ convolutional kernel followed by max pool and batch normalization. In the first, second, third, and fourth layer we have 32, 64, 128, and 256 kernels respectively following by a final layer for the output. 

\begin{figure}[!h]
  \centering
  \includegraphics[scale=.425]{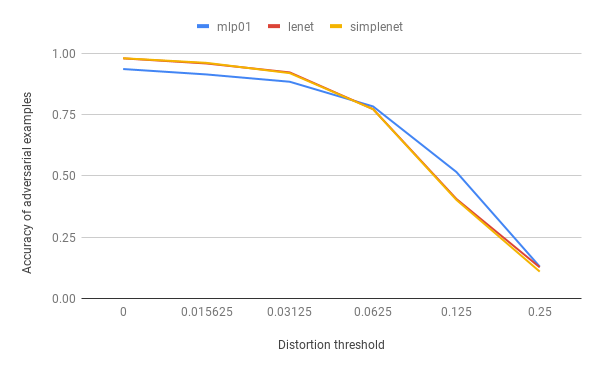} 
  \caption{Accuracy of adversarial examples for different distortion thresholds \label{cnn3}}
\end{figure}

In Figure~\ref{cnn3} we see that our model has a comparable accuracy to the convolutional models on clean test data and low distortion thresholds. However, when we cross 0.03125 then MLP01 has the highest accuracy. At threshold 0.125 it is about 11\% higher than both LeNet and SimpleNet500. 
 
\section{Discussion}
Binarized neural networks \cite{galloway2017attacking,courbariaux2016binarized,rastegari2016xnor} have weights and activations constrained to be near +1 and -1 (or 1 and 0) whereas our model weights are real numbers. The purpose of those networks is efficiency as opposed to robustness. Indeed we see in recent work that binarized networks offer marginal improvements in robustness to substitute model black box robustness on MNIST and none in CIFAR10 (see Tables 4 and 5 in \cite{galloway2017attacking} and Table 8 in \cite{panda2019discretization}). 

We make the same observations here: BNN has similar distortions to MLP and similar accuracies on adversarial examples. Both BNN and MLP01 have sign activations yet MLP01 has higher distortions and higher adversarial accuracies. Perhaps this has to do with the optimization method. BNNs are trained with an approximation to the sign activation that is differentiable whereas we train with direct coordinate descent. 

In separate work we study transferability between our 01 loss network and the standard sigmoid activated cross-entropy loss networks \cite{xue2020transferability}. There we show a lack of transferability between convex and 01 loss models in white box attacks and that both convex and 01 loss substitute model black box attacks are ineffective on our 01 loss network. However, in the work here we focus on the distortion thresholds of adversarial and Gaussian noise examples. 

Interestingly the adversarial accuracy of our network is on-par with simple convolutional models that have the powerful advantage of convolutions. As future work 01 loss convolutions may be a promising avenue to obtain models with high clean test accuracy and high adversarial accuracy as well.

\section{Conclusion}
We show that our 01 loss neural network can correctly classify images with a higher distortion than both the sigmoid activated cross-entropy loss network and binarized neural networks.  

\bibliography{my_bib}
\bibliographystyle{unsrt}

\end{document}